\documentclass[conference]{IEEEtran}
\IEEEoverridecommandlockouts
\usepackage{cite}
\usepackage{amsmath,amssymb,amsfonts}
\usepackage{algorithmic}
\usepackage{graphicx}
\usepackage{textcomp}
\usepackage{xcolor}
\usepackage{array,multirow}
\usepackage{tikz}
\usepackage[pagebackref,breaklinks,colorlinks]{hyperref}

\def\BibTeX{{\rm B\kern-.05em{\sc i\kern-.025em b}\kern-.08em
    T\kern-.1667em\lower.7ex\hbox{E}\kern-.125emX}}

\newcommand\copyrighttext{%
	\footnotesize \copyright~2023 IEEE. Personal use of this material is permitted. Permission from IEEE must be obtained for all other uses, in any current or future media, including reprinting/republishing this material for advertising or promotional purposes,creating new collective works, for resale or redistribution to servers or lists, or reuse of any copyrighted component of this work in other works.}
\newcommand\copyrightnotice{%
	\begin{tikzpicture}[remember picture,overlay]
	\node[anchor=south,yshift=10pt] at (current page.south) {\fbox{\parbox{\dimexpr\textwidth-\fboxsep-\fboxrule\relax}{\copyrighttext}}};
	\end{tikzpicture}%
}

\begin{document}

\title{GIT: Detecting Uncertainty, Out-Of-Distribution and Adversarial Samples using Gradients and Invariance Transformations\\}

\author{\IEEEauthorblockN{Julia Lust}
\IEEEauthorblockA{\textit{Robert Bosch GmbH, Stuttgart, Germany} \\
\textit{University of L\"ubeck, L\"ubeck, Germany} \\
juliarebecca.lust@de.bosch.com}
\and
\IEEEauthorblockN{Alexandru P. Condurache}
\IEEEauthorblockA{\textit{Robert Bosch GmbH, Stuttgart, Germany} \\
\textit{University of L\"ubeck, L\"ubeck, Germany} \\
alexandrupaul.condurache@de.bosch.com}
}

\maketitle

\copyrightnotice

\begin{abstract}
	Deep neural networks tend to make overconfident predictions and often require additional detectors for misclassifications, particularly for safety-critical applications. Existing detection methods usually only focus on adversarial attacks or out-of-distribution samples as reasons for false predictions. However, generalization errors occur due to diverse reasons often related to poorly learning relevant invariances. We therefore propose GIT, a holistic approach for the detection of generalization errors that combines the usage of gradient information and invariance transformations. The invariance transformations are designed to shift misclassified samples back into the generalization area of the neural network, while the gradient information measures the contradiction between the initial prediction and the corresponding inherent computations of the neural network using the transformed sample. Our experiments demonstrate the superior performance of GIT compared to the state-of-the-art on a variety of network architectures, problem setups and perturbation types.
\end{abstract}

\section{Introduction}
\label{sec:intro}
Deep Neural Networks (DNNs) have become the standard approach for a wide variety of tasks such as speech recognition and especially computer vision \cite{ciregan2012multi,ren2015towards,lecun2015deep}. Despite their success, they suffer from the tendency to make overconfident predictions. For example, the softmax score of the winning class is a bad measurement for the prediction's uncertainty \cite{hendrycks2016baseline}. However, a reliable uncertainty prediction is important, especially when DNNs are considered for safety relevant tasks, such as autonomous driving \cite{gauerhof2018structuring} or medical prognoses \cite{karabatak2009expert} where errors can have fatal consequences.

Investigating reasons for errors of a DNN is directly related to its \emph{generalization} behaviour -- the ability of a DNN to correctly classify unseen data. The generalization ability is usually evaluated using a test set independent from the training set. Both sets are sampled from the problem space typically following the same sampling distribution. The generalization ability depends on the architecture, the training procedure and especially the training data. A \emph{generalization area} can be assigned to each trained DNN as the region of the problem space in which the DNN decides reasonably correct on the input samples \cite{lust2020survey}. Among the reasons limiting the generalization area in practice are inadequate sampling of the problem-space distribution, distribution shifts or a poorly chosen model capacity. Samples outside of the generalization area will likely be misclassified and should be detected at inference time.

In this paper we focus on image classification DNNs and briefly touch upon object detection. We believe the concepts we address are applicable to many deep learning approaches. 

Currently, the topic of detecting samples outside the generalization area for image classification is covered in three distinct literature fields: \emph{Predictive Uncertainty, Adversarial Examples} and \emph{Out-of-Distribution Detection}. Each field investigates a specific reason for misclassification and proposes different methods geared towards the detection of the corresponding misclassified samples. \emph{Predictive Uncertainty} considers misclassifications that occur randomly and typically close to samples inside the generalization area \cite{blundell2015weight,gal2016dropout,shen2021real,malinin2019ensemble}. \emph{Adversarial Examples} are constructed to fool the DNN on purpose for example by shifting the image via slight changes into small pockets of limited occurrence probability during training \cite{ma2018characterizing,meng2017magnet,liao2018defense,xu2018feature,ma2019nic,lee2018simple,cohen2020detecting,lust2020gran}. \emph{Out-of-Distribution} data includes samples that are from outside the problem space, e.g. samples from another dataset \cite{liang2017enhancing,lee2018simple,Hsu_2020_CVPR,oberdiek2018classification,ren2019likelihood,yu2019unsupervised,sastry2020detecting}.

\begin{figure*}[t]
	\begin{center}		
		\begin{tikzpicture}[shorten >=1pt,->]
		\tikzstyle{unit}=[draw,shape=circle,minimum size=1.15cm]
		
		\node[rectangle,minimum width=3.cm,minimum height = 1.5cm](p) at (2.26,1.25){\includegraphics[height=8ex]{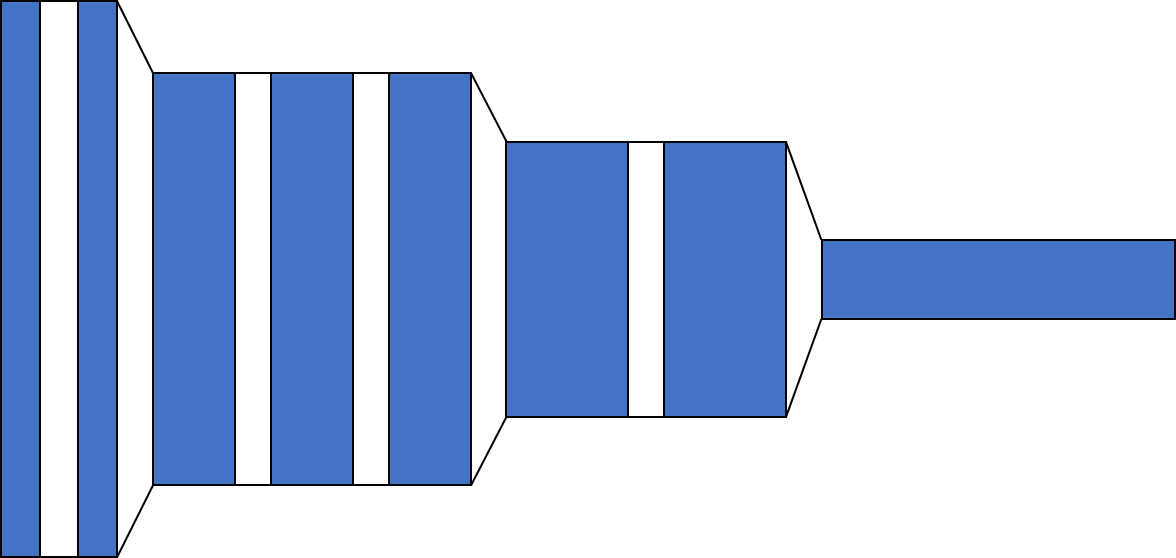}};
		\node[below of=p] (forward)  {forward pass};
		\node[above of=p] (forward)  {DNN};
		\node(x) at (-1.5,1.25){\includegraphics[height=3ex]{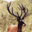}};
		\node[above of=x,yshift=-18.]  {$x_0$};
		\node(dots) at (-0.25,1){\vdots};
		
		\node(x0)at (-0.25,2.75){\includegraphics[height=3ex]{Figures/deer6.png}};
		\node(x1)at(-0.25,1.87){\includegraphics[height=3ex]{Figures/deer6.png}};
		\node(xn)at(-0.25,-0.25){\includegraphics[height=3ex]{Figures/deer6.png}};
		\draw (x)--node [above,sloped] {\small Id.}(x0.180);
		\draw (x)--node [above,sloped,yshift=-2.] {\small T$_1$}(x1);
		\draw (x)--node [above,sloped]{\small T$_N$}(xn.180);
		\draw (x0)--(p.160);
		\draw (x1)--(p.170);
		\draw (xn)--(p.200);
		
		\node(fx0)at (5,2.5){$F(x_0)$};
		\node(fx1)at(5,1.75){$F(x_1)$};
		\node(fxn)at(5,0){$F(x_N)$};
		\node(dots) at (5,1){\vdots};
		
		\draw (p.20)--(fx0.180);
		\draw (p.10)--(fx1.180);
		\draw (p.-20)--(fxn.180);
		
		\node(y)at (7.6,3.25){$y$};
		\node(lfx)at (7,2.5){$L(F(x_0),y)$};
		\node(lfx1)at(7,1.75){$L(F(x_1),y)$};
		\node(lfxn)at(7,0){$L(F(x_N),y)$};
		\node(dots) at (7,1){\vdots};
		
		\draw (fx0)--(lfx.180);
		\draw (fx1)--(lfx1.180);
		\draw (fxn)--(lfxn.180);
		
		\node[rectangle,minimum width=3.cm,minimum height = 1.5cm](dnn2) at (10.,1.25){\includegraphics[height=8ex]{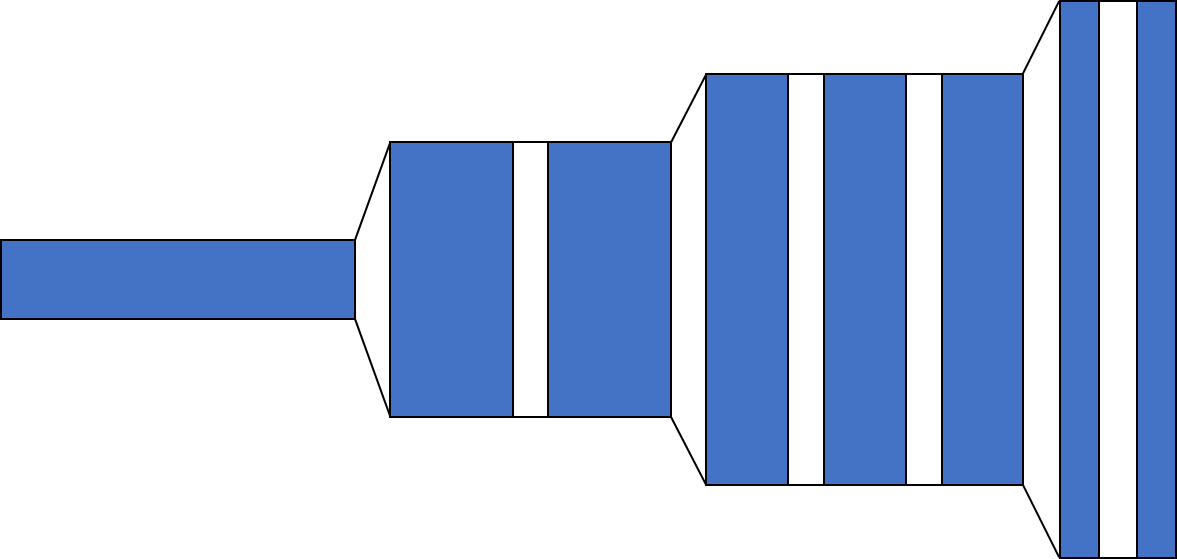}};
		\node[below of=dnn2] (forward)  {backward pass};
		\node[above of=dnn2] (forward)  {DNN};
		
		\draw (fx0.30)--(y.180);
		\draw (y.270)--(lfx.27);
		\draw (lfx.0)--(dnn2.160);
		\draw (lfx1.0)--(dnn2.170);
		\draw (lfxn.0)--(dnn2.200);
		
		\node(glfx)at(13.,2.5){\large $\frac{\partial L(F(x_0),y)}{\partial \omega}$};
		\node(glfx1)at(13.,1.75){\large $\frac{\partial L(F(x_1),y)}{\partial \omega}$};
		\node(glfxn)at(13.,0){\large $\frac{\partial L(F(x_N),y)}{\partial \omega}$};
		\node(dots) at (13.,1){\vdots};
		
		\draw (dnn2.20)--(glfx.180);
		\draw (dnn2.10)--(glfx1.180);
		\draw (dnn2.-20)--(glfxn.180);
		
		\node[rectangle,minimum width=0.5cm,minimum height = 1.5cm](single) at (15.00,1.25){\includegraphics[height=5ex]{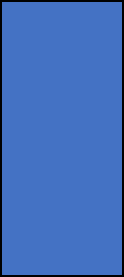}};
		\node[above of=single]   {Head};

		\draw (glfx.0)--(single);
		\draw (glfx1.0)--(single);
		\draw (glfxn.0)--(single);
		
		\node(pe) at (15.75,1.25){$p$};
		
		\draw (single) --(pe);
		\end{tikzpicture}
	\end{center}
	\caption{GIT architecture. GIT detects input samples outside the generalization area of the DNN during inference time.}
	\label{Fig:GIT}
\end{figure*}

Misclassifications occur when data points undergo various perturbations that push them outside of the generalization area. The perturbation moves a data sample along invariance directions up to the point where it surpasses the amount of invariance that is captured by the model. We believe that current literature lacks a method and an inference time evaluation procedure that combines the following objectives:
\begin{enumerate}
	\item[O1] \textbf{Detecting misclassifications caused by (real-world) relevant perturbation types}.  Most current literature only considers adversarial perturbations which are not as real-world relevant as e.g. the corruptions proposed by Hendrycks et al. in their ImageNet-C perturbation collection \cite{hendrycks2019benchmarking}.
	\item[O2] \textbf{Considering perturbed but correctly classified data.} If a classifier is invariant to perturbations up to a certain amplitude, then only some samples would be wrongly classified. Perturbed samples not leading to a wrong prediction by the DNN should not be detected as a misclassification but accepted as correctly classified. In current set-ups these samples are either ignored or even considered in the class of misclassifications.
	\item[O3] \textbf{One detection method that generalizes to all reasons of misclassifications.} Many methods are trained and tested for each corruption type separately. However, in real-world applications several different types of corruptions may be relevant for the generalization behavior. Therefore, a setup that properly evaluates the generalization ability of the detection method is necessary.
\end{enumerate}
Some of these objectives are met by the methods \textit{Mahalanobis} \cite{lee2018simple} and \textit{GraN} \cite{lust2021efficient}. While Mahalanobis is evaluated on adversarial as well as out-of-distribution data, the authors did neither consider perturbed data that is still correctly classified nor real-world relevant perturbations. GraN, on the other hand considers a large amount of different perturbation types and furthermore includes some correctly classified perturbed samples in its evaluation. However, the method has not been tested for its generalization ability between different perturbations. We found that both methods have weaknesses when tested on a setup that meets all the above objectives. 

Therefore, we introduce \textbf{GIT} (Fig.~\ref{Fig:GIT}), which is designed to consider several reasons for misclassification using \textbf{G}radient features and multiple \textbf{I}nvariance \textbf{T}ransformations. The latter are based on prior knowledge about key invariances that need to be properly handled by the DNN for a correct decision. They are thought to shift samples outside the generalization area back into it, while keeping samples already inside the generalization area within it. For shifted samples this increases the contradiction between the DNN and the prediction, e.g. if many paths of the network characterised by the corresponding weights would hint towards another output of the network for the transformed input. This contradiction can be measured by the gradient of the weights computed for the loss function comparing the original and the transformed output. 

We evaluate GIT on two classification network architectures (ResNet and DenseNet) and on four dataset (SVHN, CIFAR-10, CIFAR-100 and ImageNet). According to our objectives, we use eleven different perturbation types (O1), include perturbed data in the class of correctly classified images (O2) and evaluate GIT when it only has access to one of the perturbation types during training and has to generalize to the others (O3). Additionally, we perform experiments for object detection using the KITTI dataset and a Single-Shot-Detector network to further test the adaptability of GIT. GIT achieves close to state-of-the-art performance for out-of-distribution tasks and significantly outperforms the baseline methods for more general real-world and adversarial corruptions. Additionally, it is able to generalize among perturbations and it can also be applied successfully to object detection.

\section{Related Work}
\label{sec:related}
There are currently three literature fields that address generalization failures at inference time: Adversarial example detection, out-of-distribution detection and predictive uncertainty. Each field covers one specific reason for misclassifications and only few works deal with connections across those literature fields. However, when having a closer look into the methods of these fields they all can be split into four main categories \cite{lust2020survey}: \textbf{generative, inconsistency, ensemble} and \textbf{metric} based approaches.

\textbf{Generative methods} are typically based on an Autoencoder or a Generative Adversarial Network that is trained to shift images into the direction of the training distribution. For adversarial images the goal is to remove the adversarial perturbation and gain an image that can be correctly classified \cite{meng2017magnet,liao2018defense}. In the case of out-of-distribution, a huge difference between the input and the output image of the generative method hints towards a distance to the training distribution \cite{ren2019likelihood}.

Similarly, \textbf{inconsistency methods} based approaches expect the output of the DNN for a misclassified image to be more sensitive to small changes in the image. In comparison to generative methods they use rather simple transformations. For out-of-distribution detection, ODIN is a well-known approach \cite{liang2017enhancing}. Its transformation is based on a gradient descent procedure with step-size $\epsilon$ performed on the original image $x$ maximising the loss function $L$ computed for the output of the network $F(x)$ and the predicted class $y$

\begin{equation}
\label{odin}
\tilde{x}=x-\epsilon \cdot \text{sign}(-\nabla_x L(x,y))\,.
\end{equation} 

Hsu et al. introduced a more sophisticated version of ODIN \cite{Hsu_2020_CVPR}. Unfortunately, existing inconsistency methods do not work well for some adversarial attacks \cite{xu2018feature,ma2019nic}.

\textbf{Ensemble methods} are based on an ensemble of DNNs. The more the outputs of the DNNs differ for the same input, the more the image is expected to be misclassified. Well known examples are Monte-Carlo-Dropout \cite{gal2016dropout} and methods building upon Bayesian Neural Networks \cite{blundell2015weight}. Furthermore, there are ensembles in which each network is trained on slightly different training sets or output requirements \cite{yu2019unsupervised,vyas2018out}. Recently, distillation methods based on ensemble results are gaining more attention. The ensemble of models allow a single DNN to explicitly model a distribution over the outputs \cite{malinin2019ensemble,shen2021real}. Ensemble methods are mainly used in the category of Predictive Uncertainty and out-of-distribution detection.

\textbf{Metric methods} use stochastic principles to evaluate whether the current input sample is behaving similarly to correctly classified input samples investigated during training using \textit{activation} or \textit{gradient} information. Earlier, such methods were typically based on the activation outputs of the network. A well-known Adversarial Examples detection approach is LID \cite{ma2018characterizing}. It is based on a Local Intrinsic Dimensionality score, a weighted distance based on k-nearest neighbours from the training set. The method Mahalanobis works similar to LID and is based on the Mahalanobis distance $M(x)_l$ computed for each layer $l$ of the DNN \cite{lee2018simple}. The Mahalanobis distance is computed for each layer-output $f(x)_l$ of the test sample $x$ and the closest class-conditional Gaussian distribution defined by the mean of all layer-outputs $\mu_y$ and the layer-output covariance $\Sigma_y$ for training samples of the class $y$
\begin{equation}
M(x)_l=- (f(x)_l - \mu_y) \Sigma_y ^ {-1} (f(x)_l-\mu_y)\,.
\end{equation}
The Mahalanobis distance is combined with a gradient shift procedure similar to that described in \eqref{odin}. Another new approach for adversarial detection is using additional procedures such as classification networks consisting of one fully connected softmax layer on top of each activation layer \cite{ma2019nic} or more sophisticated k-nearest neighbour procedures in combination with influence functions \cite{cohen2020detecting}. Other metric based methods in this field are dominated by computationally expensive layer-wise higher-order Gram matrices \cite{sastry2020detecting} or additional residual flow procedures \cite{zisselman2020deep}. Currently \emph{gradient} based methods are gaining importance in which the gradient of the network regarding a loss function computed on the predicted class and the corresponding softmax output is serving as the features for the detector \cite{oberdiek2018classification, lust2020gran, lee2020gradients, lust2021efficient, huang2021importance}. Recently GraN \cite{lust2020gran, lust2021efficient}, that combines gradients with a gaussian smoothing, outperformed the state-of-the art method Mahalanobis \cite{lee2018simple} in the combined area of Adversarial Examples and Out-of-Distribution detection. GradNorm \cite{huang2021importance} stated state-of-the-art performance in the detection of Out-of-Distribution samples. GradNorm uses  an uniform vector instead of the predicted class vector for gradient computation and only consider the gradients of the last layer.

Our method GIT combines the fields of metric and inconsistency methods by using invariance transformations and gradient information. Mahalanobis \cite{lee2018simple} and GraN \cite{lust2021efficient} are the only methods that have been evaluated on more than one reason for misclassification. Due to their state-of-the-art performance in this field they will serve as baseline methods in our experiments. As an additional baseline we evaluated GradNorm, since it is similarly to GIT based on gradients and currently achieves state-of-the-art in the field of out-of-distribution detection.

\section{Method}
\label{sec:method}

GIT detects whether an input is misclassified during inference time of a DNN. The architecture of GIT is visualised in Fig.~\ref{Fig:GIT}. GIT consists of three components: Invariance transformations shift misclassified data samples $x$ back into the direction of the generalization area while not affecting correctly classified samples, the feature extraction based on gradient information and the head that combines the features into the output $p \in [0,1]$ stating whether the input $x$ is misclassified. A GIT stream is defined by one transformation in combination with the corresponding gradient based feature extraction. The head fuses all streams.

The intuition behind GIT is that three possible cases can occur when distinguishing a data point outside, from a data point inside of the generalization area:
\begin{enumerate}
	\item Data from inside the generalization area is modified by the transformations in order to eliminate variance along directions of invariance already captured by the model. Therefore this data remains in the generalization area and correctly classified data does not lead to high gradients.
	\item Data from outside the generalization area can be shifted inside the generalization area by at least one transformation. In this case the transformation eliminates the variance along directions of invariances that are not captured in the model. This leads to high gradients used as features in the detection head of GIT. The data point is detected as outside of the generalization area.
	\item If the data point is outside of the generalization area but a single stream is not able to provide meaningful gradient features, the multi-stream approach allows several other streams to contribute meaningful gradients. These can be leveraged by the detection head of the classification chain, leading to a successful detection.
\end{enumerate}

\subsection{Invariance Transformations}
Generalization errors are caused by bad learning of invariances. Each stream of the multistream architecture is based on an invariance transformation. The transformations cover prior knowledge on invariances that need to be captured by the classifier in order to correctly classify samples \cite{rath2020boosting}. Depending on the amount of prior knowledge concerning relevant invariances, the number and the concrete transformations can be adjusted. There is one stream per invariance transformation plus an additional identity stream as discussed above.  In this paper we propose three invariance streams based on filter applications dealing with global invariances and one stream based on an autoencoder thought to cover local invariances. Each stream $T_i$ generates a transformed image 
\begin{equation}
x_i = T_i(x_0)
\end{equation}
from an input image $x_0$.

\textbf{Gaussian} filters use a two dimensional symmetric kernel derived from a Gaussian distribution $G(u,v)$ approximated for discrete pixel values \cite{haralick1992computer}. The resulting discrete values build a filter mask. This filter was originally used in the smoothing step of GraN \cite{lust2020gran,lust2021efficient}. It is supposed to eliminate classification errors due to additive noise. Furthermore, to a certain extent it also eliminates high-frequency image content that often is responsible for spurious correlations being learned during training.

\textbf{Wiener} filtering eliminates classification errors due to poorly captured invariance to sensor noise and blur caused by for example poor optics. It is minimizing the mean squared error between a noisy image and the filtered image under the assumption of a known, stationary noise and frequency response of the imaging system \cite{haralick1992computer}. The noise level is a hyper-parameter defining the final filter.

\textbf{Median} filtering eliminates classification errors due to poorly captured invariance to salt-and-pepper noise, e.g. caused by corrupt pixels. A Median filter selects the median value from all pixels in the appropriate square neighbourhood around the target pixel. The Median filter only depends on the hyper-parameter defining the size of the filter window.

\textbf{Autoencoders} \cite{sakurada2014anomaly} are supposed to eliminate classification errors due to poorly captured unspecified in-distribution invariances. They are trained using a loss function comparing the input to the output image for images of the training set of the method under test, i.e., the classifier whose generalization area is analysed.
\subsection{Gradient Information}
For each stream $i \in \lbrace 1,\dots,n \rbrace$ the gradients measure the contradiction between the current prediction $y$ and the output of the transformed sample $F(x_i)$ within the network. For the computation of the gradients the classification DNN performs a foward pass for the original image $x$, and the transformed image $x_i$. For the original image $x$ the predicted class $y$ is derived as the index of the largest value of the output $F(x_0)$ of the network for the original image $x_0$. The output $F(x_i)$ of the transformed image $x_i$ and the predicted class $y$ as one-hot vector are compared using the cross entropy loss function $L(\cdot,\cdot)$. In the next step, the gradient 
for the network weights $\omega$ is computed by a backward pass of the DNN.

The large vector of gradients is reduced to a smaller set of features applying a layer-wise average pooling
\begin{equation}
\label{norm}
\dfrac{\partial L(F(x_i),y)}{\partial \omega}~\mapsto~ {\begin{pmatrix}||\dfrac{\partial L(F(x_i),y)}{\partial \omega_1}||_1\\\vdots \\||\dfrac{\partial L(F(x_i),y)}{\partial \omega_L}||_1\end{pmatrix}}.
\end{equation}
For each layer $l \in \lbrace1,\dots,L\rbrace$ of the DNN, the gradients regarding the layer's weights $\omega_l$ are replaced by their $L_1$ norm. This results in a feature vector of size $L$ which reflects the number of layers in the DNN.

\subsection{Head}
For each stream the feature vector is processed by a logistic regression network. The outputs of all streams are then combined by another logistic regression network to one single value $p$. \\
\textbf{Training:} Training and validation data for the misclassification detection task (i.e., analysing the generalization area) can be gathered from correctly and misclassified samples of the classification dataset. Each logistic regression stream is trained and hyper-parameter optimised individually. Then, the logistic regression network combining the individual streams is trained on the predicted outputs.

\section{Evaluation}
\label{sec:exp}

We evaluate GIT on several problem setups corresponding to the classification datasets CIFAR-10, CIFAR-100 \cite{krizhevsky2009learning}, SVHN \cite{netzer2011reading} and ImageNet \cite{deng2009imagenet} and on two popular models: DenseNet \cite{huang2017densely} and a ResNet \cite{he2016deep}. The DNNs and their training procedure are adopted from reference \cite{lee2018simple} and \cite{huang2021importance} for ImageNet (see Appendix for details). For each classification problem, we use the corresponding training set to train the models and build all perturbation setups based on the test set. Their generation is described in the next section. Each perturbation setup is split randomly into 80\% training data, 10\% validation data and 10\% test data. As explained in Section 2 we use Mahalanobis, GraN and GradNorm as baseline methods. The trainable parts of Mahalanobis, GraN, GradNorm and GIT (own) are trained on the perturbation-setup training data, validated on the validation data and tested on the test set for each setup. The range of possible hyper-parameters for Mahalanobis are adopted from reference \cite{lee2018simple}. For GraN and GIT the standard deviation $\sigma$ for the Gaussian smoothing is chosen from $\sigma \in \lbrace 0.1,0.2.,\dots, 1.0 \rbrace $. For GIT the Median filter size is chosen from $\lbrace 2\times 2, 3\times 3,\dots, 10 \times 10\rbrace$ and the noise level for the Wiener filter from $\lbrace 0.01,0.02.,\dots, 0.10 \rbrace $. We trained the autoencoder on the problem-setup training set (see Appendix for implementation details). As evaluation score we use the area under the receiver operating characteristic curve (AUROC),which plots the true positive rate (TPR) against the false positive rate (FPR).

\subsection{Perturbation setups}
\label{Sec:Pert}
A sample is misclassified by a DNN if it is not within the generalization area of the DNN. Possible reasons for a sample to be outside the generalization area can be split into three main categories: Predictive Uncertainty, Adversarial Examples and Out-of-Distribution Detection. According to those categories we build eleven perturbation setups.

\begin{table*}[t]
	\caption{AUROC score for the detection methods Mahalanobis, GraN, GradNorm and GIT (own). \emph{(seen)} marks the perturbation set used for training the heads of the trainable detectors.}	
	\begin{center}
		\begin{tabular}{l l lc  c c c}
			\hline
			\multirow{3}{*}{\rotatebox[origin=c]{90}{\textbf{Model}}}&&\multicolumn{1}{r}{\textbf{Data}} & \textbf{CIFAR-10}&\textbf{CIFAR-100}&\textbf{SVHN} & \textbf{ImageNet} \\
			&\multicolumn{2}{r}{\textbf{Detector}}& \multicolumn{4}{c}{\textbf{Mahalanobis / GraN / GradNorm / GIT (own)}}\\

			&\multicolumn{2}{l}{\textbf{Perturbation}}& & & &\\

			\hline
			\multirow{12}{*}{\rotatebox[origin=c]{90}{ResNet}}&\multirow{4}{*}{\rotatebox[origin=c]{90}{P.Unc.}}&Original &77.15/80.28/70.08/\textbf{89.42}~&~\textbf{84.22}/79.72/59.72/81.26~& 86.07/86.53/50.85/\textbf{87.01}&48.79/81.05/64.70/\textbf{81.53}\\
			&&Gaussian &71.85/72.78/63.51/\textbf{94.09}& 80.09/84.48/61.52/\textbf{92.15}& 79.26/90.18/62.07/\textbf{96.45}&60.79/84.44/73.24/\textbf{84.96}\\
			&&Shot &   71.98/72.39/62.14/\textbf{92.94}& 81.26/84.09/60.29/\textbf{91.43} &  76.11/90.54/63.12/\textbf{96.38}&59.77/84.57/72.95/\textbf{86.16}\\
			&&Impulse & 83.90/84.36/67.58/\textbf{95.75} & 82.08/81.76/63.93/\textbf{89.96}&78.80/86.93/63.64/\textbf{97.18}&58.99/84.67/71.86/\textbf{85.30}\\
			
			\cline{2-7}
			&\multirow{4}{*}{\rotatebox[origin=c]{90}{Adv.}}&FGSM(seen) &  77.47/84.13/56.91/\textbf{96.57}& 83.36/84.31/64.58/\textbf{92.23}& 74.18/75.87/55.91/\textbf{91.78}&57.92/81.58/72.36/\textbf{84.64}\\
			&&BIM  & 80.37/88.67/61.22/\textbf{97.90}& 83.06/85.40/70.66/\textbf{92.07}&  81.96/56.64/64.67/\textbf{95.35}&58.80/84.26/72.90/\textbf{85.10}\\
			&&DeepFool& 45.10/81.59/62.27/\textbf{94.43}& 41.87/70.58/60.19/\textbf{80.27}& 36.93/72.85/58.38/\textbf{93.40}&---\\
			&&CWL2 &   82.56/91.21/67.96/\textbf{98.68}& 88.25/88.74/52.77/\textbf{93.20}& 85.63/88.51/52.43/\textbf{98.12} &52.02/90.40/63.19/\textbf{92.17}\\
			\cline{2-7}
			&\multirow{3}{*}{\rotatebox[origin=c]{90}{OOD}}&SVHN/C/Nat   &\textbf{96.77}/65.23/57.31/76.42& \textbf{78.71}/69.73/53.32/67.79&\textbf{95.96}/93.45/57.15/93.04&42.24/91.77/92.61/\textbf{92.92}\\
			&&ImgN/Places & 74.04/78.88/71.58/\textbf{88.03}&63.55/78.02/57.76/\textbf{79.50}&\textbf{96.14}/95.42/57.50/93.84&47.27/89.61/85.62/\textbf{90.46}\\
			&&LSUN/SUN &  80.22/83.45/72.03/\textbf{88.25}& 69.61/79.18/55.24/\textbf{79.52}& \textbf{95.56}/94.60/55.10/92.94&41.17/88.84/\textbf{94.24}/89.86\\
			\hline
			\multirow{12}{*}{\rotatebox[origin=c]{90}{DenseNet}}&\multirow{4}{*}{\rotatebox[origin=c]{90}{P.Unc.}}&Original &49.87/79.22/78.29/\textbf{87.22}~&~59.53/\textbf{81.81}/67.52/79.59~&\multicolumn{1}{r}{79.99/83.46/54.50/\textbf{85.89}}&50.63/92.87/56.83/\textbf{94.63}\\
			&&Gaussian &56.08/83.21/72.24/\textbf{96.22} & 50.32/86.75/70.77/\textbf{90.05}& 70.70/88.95/58.94/\textbf{97.27}&53.37/82.93/70.45/\textbf{83.49}\\
			&&Shot & 54.38/80.04/75.04/\textbf{95.09}& 48.02/86.74/71.00/\textbf{89.42}& 69.64/87.45/54.06/\textbf{96.80}&54.24/83.00/71.43/\textbf{83.78}\\
			&&Impulse &54.31/85.26/66.33/\textbf{96.00}& 48.53/82.77/66.83/\textbf{89.29} & 71.32/84.48/58.90/\textbf{96.32}&49.17/83.60/69.89/\textbf{83.89}\\
			
			\cline{2-7}
			&\multirow{4}{*}{\rotatebox[origin=c]{90}{Adv.}}&FGSM(seen) & 53.82/88.55/55.39/\textbf{97.50}& 53.39/86.82/62.10/\textbf{91.50}& 73.75/81.17/51.64/\textbf{93.09}& 51.75/\textbf{84.03}/69.94/83.92 \\
			&&BIM & 56.91/91.91/54.97/\textbf{98.42}& 52.56/87.94/63.63/\textbf{91.79}&  80.55/76.93/79.02/\textbf{96.52}&52.06/83.28/70.22/\textbf{83.38}\\
			&&DeepFool &52.97/82.84/51.35/\textbf{91.82}& 55.96/76.41/56.48/\textbf{80.60}& 61.00/91.73/65.06/\textbf{94.83}&---\\
			&&CWL2 &  56.70/93.67/64.00/\textbf{98.76}&  51.37/92.46/68.38/\textbf{93.06}& 77.21/95.90/63.50/\textbf{98.57}&51.52/96.10/52.33/\textbf{96.80} \\
			\cline{2-7}
			&\multirow{3}{*}{\rotatebox[origin=c]{90}{OOD}}&SVHN/C/Nat &78.82/70.38/\textbf{84.75}/73.87&19.61/77.79/\textbf{92.82}/71.80&91.89/92.50/53.30/\textbf{93.72} & 50.75/95.74/93.78/\textbf{96.38} \\
			&&ImgN/Places & 10.69/83.10/\textbf{95.29}/83.66&33.10/\textbf{74.89}/64.54/57.76&89.94/\textbf{95.47}/68.31/94.60&58.05/94.46/85.41/\textbf{94.72}\\
			&&LSUN/SUN & 11.69/85.86/\textbf{96.96}/85.37 &  39.48/\textbf{78.53}/58.99/58.03&92.10/\textbf{94.59}/73.06/93.69&43.25/95.34/91.48/\textbf{95.78}\\
			\cline{1-7}
		\end{tabular}		
	\end{center}
	\label{Tab:Final}
\end{table*}

\textbf{Predictive Uncertainty (P.Unc.):} The \textbf{original} perturbation setup is build by the samples that have been misclassified by the classification network and the same amount of randomly chosen correctly classified images of the problem-setup test dataset. For the \textbf{Gaussian, Shot} and \textbf{Impulse} setup the test data is corrupted with the corresponding noise. The noise level is adapted such that half of the original data is misclassified and half of the data is still correctly classified by the DNN.\\
\emph{Explanation:} These setups cover \emph{normal} misclassifications and misclassifications caused by corruptions e.g. due to the optical path and sensor setup. Sensor related corruptions are relevant in most computer vision tasks. Similar to Lust and Condurache \cite{lust2021efficient} we therefore based these setups on the sensor related corruption types introduced by Henrycks et al. \cite{hendrycks2019benchmarking}: \emph{Gaussian} noise is based on a normal distribution and can appear in low-lighting conditions. \emph{Shot} noise is generated using a poisson distribtuion and simulates electronic noise caused by the discrete nature of light. \emph{Impulse} noise is similar to salt-and-pepper noise for black and white images and can be used to simulate bit errors.

\textbf{Adversarial Examples (Adv.):} To generate Adversarial Examples four different attack methods are used: \textbf{FGSM} \cite{goodfellow2014explaining}, \textbf{BIM} \cite{kurakin2016adversarial}, \textbf{Deepfool} \cite{papernot2016limitations} and \textbf{CWL2} \cite{carlini2017towards}. The generation of the Adversarial Examples is adopted from Lee et al. \cite{lee2018simple}. Each attack is applied to the problem-setup test images with a corruption level leading to a misclassification in 50\% of the images. \\
\emph{Explanation:} Adversarial Examples are artificially generated samples that are constructed to fool a network into making a false decision. Usually, Adversarial example methods shift a correctly classified, original image such that the predicted class changes while the difference is constructed to be as small as possible. Common Adversarial Examples detection setups only consider adversarially perturbed images that lead to a misclassification. We consider both, adversarial images leading to a misclassification in the set of wrongly classified images and images not leading to a misclassification in the set of correctly classified images.

\textbf{Out-Of-Distribution (OOD):} In the Out-of-Distribution detection setups we used datasets different to the current training data as Out-of-Distribution data. If the current training data is CIFAR-10/100 then\textbf{ SVHN} \cite{netzer2011reading}, \textbf{TinyImageNet (ImgN)} \cite{le2015tiny} and \textbf{LSUN} \cite{yu2015lsun} are used as Out-of-Distribution data, similar for SVHN where \textbf{CIFAR-10 (C)} \cite{krizhevsky2009learning} replaces SVHN. For ImageNet the corresponding large scale datasets \textbf{iNaturalist (Nat)} \cite{van2018inaturalist} \textbf{Places} \cite{zhou2017places} and \textbf{SUN} \cite{xiao2010sun} are used. Each Out-of-Distribution setup is built from one of the Out-of-Distribution dataset and correctly classified images from the problem-setup test data. \\
\emph{Explanation:} This procedure is standard for Out-of-Distribution detection. It simulates the occurrence of data not present in the training distribution.

%
%

\subsection{Results and Ablation Studies}
In the following, we evaluate the performance of GIT in comparison to other methods, provide two ablation studies and show how to adapt GIT for object detection.

\begin{table*}[t]
	\caption{AUROC score of GIT depending on the combination of the invariance transformation streams for DenseNet and CIFAR-10.}
	\begin{center}
		\begin{tabular}{cl c c c c c }
			\hline
			&\multicolumn{1}{r}{\textbf{Streams}} &\textbf{Gaus.}  & \textbf{AE}& \textbf{Gaus.+AE} & \textbf{Gaus.+AE+Median}& \textbf{Gaus.+AE+Median+Wiener}\\
			\multicolumn{2}{l}{\textbf{Perturbation}} & \\
			\hline
			\multirow{4}{*}{\rotatebox[origin=c]{90}{P.Unc.}}&Original  & 79.22&80.73& 83.65&86.32&\textbf{87.22}\\
			&Gaussian  & 83.21&93.25& 93.64&95.01&\textbf{96.22} \\
			&Shot  &80.04&92.92& 92.43&93.59&\textbf{95.09} \\
			&Impulse & 85.26 &90.72& 92.43&\textbf{97.31}& 96.00\\
			\hline
			\multirow{4}{*}{\rotatebox[origin=c]{90}{Adv.}}&FGSM (seen) & 88.55 &95.72& 95.69&97.21& \textbf{97.50}\\
			&BIM & 91.91 &96.27& 96.78&98.24& \textbf{98.42} \\
			&DeepFool &82.84&88.34&89.32&91.27& \textbf{91.82}\\
			&CWL2  & 93.67&96.38& 97.30&98.54& \textbf{98.76} \\
			\hline
			\multirow{3}{*}{\rotatebox[origin=c]{90}{OOD}}&SVHN  &70.38&70.84& 69.85&\textbf{75.09}&73.87  \\
			&ImageNet &83.10 & 74.96 &78.56& 82.72&\textbf{83.66}\\
			&LSUN & 85.86 &76.86& 81.05& \textbf{86.15} & 85.37\\
			\hline
			
		\end{tabular}
		
	\end{center}
	\label{Tab:Streams}
\end{table*}
\subsubsection{Comparing GIT to other Methods}
We evaluated the performance of Mahalanobis, GraN, GradNorm and our detection method GIT (Tab.~\ref{Tab:Final}). For each dataset, model and perturbation combination the detection methods were trained only on the corresponding FGSM perturbation setup. We report the AUROC scores for each combination and method, the resulting TNR at TPR 95\% evaluation is in the Appendix. Since the adversarial method DeepFool relies on calculating the gradients for each possible class per sample, it is computationally infeasible to apply it to the large scale dataset ImageNet with its 1000 classes. Furthermore, since simple autoencoders do not work well for large scale datasets we only consider the gaussian, median and wiener transformations in the case of ImageNet.  

GIT significantly outperforms GradNorm, GraN and Mahalanobis among a wide variety of set-ups, except for some Out-of-Distribution setups, showcasing its generalization ability among a multitude of perturbations. The distance-based method Mahalanobis achieves a good AUROC score only in the Out-of-Distribution setup. Its performance is particularly poor for DenseNet that has many skip connections allowing information flow between all regions of the network. The Adversarial Example and Predictive Uncertainty setups are build such that the data points lie close to the classification boundary of the DNN. Correctly and incorrectly classified samples are similarly far away from the original data distribution. Distance-based methods such as Mahalanobis are good in detecting samples far away from the original distribution as in the case of Out-of-Distribution setups. However, they are not able to consider whether the samples are on the correct side of the decision boundary. On the contrary, gradient based methods consider the contradiction within the network to the predicted class via the gradient. They are therefore more capable of detecting samples outside but close to the generalization area.

The performance of GIT for the Out-of-Distribution setups is mixed. In case of OOD the invariance transformations are unable to shift the data point back inside the generalization area which impedes the detection of such kind of data. This effect can be compensated when adding Out-of-Distribution data during the training of GIT. In our ablation study (Sec.~\ref{Abl.Perturbations}, Tab.~\ref{Tab:Data}) GIT could be significantly improved for Out-of-Distribution detection when increasing the variabilty in the perturbation-setup training data by adding Out-of-Distribution and Predictive Uncertainty training samples besides FGSM data. 
GradNorm uses only the gradients of the last layer as feature input. This makes the performance more dependent on the used data and classification DNN. We considered this by adapting GradNorm using the gradients of all layers which improved its performance but could not outperform GIT (see Appendix). GradNorm uses an uniform vector as target while GraN and GIT use the predicted class one-hot vector. The important difference is that for GradNorm only one of the two inputs of the loss function depends on the actual input. The integration of invariance transformations in order to compare the output of the original and a transformed image is therefore not possible. GraN and GIT can use this idea, while GIT additionally has the advantage of the multistream architecture which enables it to further generalize by considering which gradient of which transformation stream carries the most relevant information for the specific unknown perturbation. This advantage allows GIT to significantly outperform the other methods in most problem setups.

\begin{table}[t]
	\caption{AUROC score depending on the perturbation data seen during training of GIT for DenseNet and CIFAR-10.}
	\begin{center}
		\begin{tabular}{cl c c c c c }
			\hline
			&\multicolumn{1}{r}{\textbf{Seen}} &\textbf{Original}  & \textbf{FGSM}& \textbf{All Perturbations}  \\
			\multicolumn{2}{l}{\textbf{Perturbation}} & \\
			\hline
			\multirow{4}{*}{\rotatebox[origin=c]{90}{P.Unc.}}&Original  & \textbf{87.74} & 87.22&87.27\\
			&Gaussian  & 61.30 & 96.22&\textbf{96.98}\\
			&Shot  &61.83 & 95.09&\textbf{95.96}\\
			&Impulse  & 59.06& 96.00&\textbf{96.81}\\
			\hline
			\multirow{4}{*}{\rotatebox[origin=c]{90}{Adv.}}&FGSM & 52.73 & \textbf{97.50}&96.89\\
			&BIM & 46.93 & \textbf{98.42}&98.16\\
			&DeepFool & 17.02& \textbf{91.82}&91.59\\
			&CWL2  & 66.67 & \textbf{98.76}&\textbf{98.76}\\
			\hline
			\multirow{3}{*}{\rotatebox[origin=c]{90}{OOD}}&SVHN  & 46.67  & 73.87& \textbf{81.32} \\
			&ImageNet & 85.82  & 83.66 &\textbf{88.85}\\
			&LSUN & 87.34 & 85.37&\textbf{91.05}\\
			\hline
		\end{tabular}
		
	\end{center}
	\label{Tab:Data}
\end{table}

\begin{table*}[t]
	\caption{mAP score for object detection on KITTI data comparing the baseline (=SSD only), Monte-Carlo Dropout, Deep Ensembles and five combinations of invariance transformation streams for GIT.}
	\begin{center}
		\begin{tabular}{ l|c c c  |c c c c c}
			\hline
			\multicolumn{1}{r|}{\textbf{Detector}} &  \textbf{Baseline}  &\textbf{MC-Drop.} & \textbf{DeepEns.}  &  \multicolumn{5}{c}{\textbf{GIT adaptations (own)}}  \\
			\multicolumn{1}{r|}{Features} &  output   &outputs & outputs &output& gradient & gradient & gradient& gradient\\
			\multicolumn{1}{r|}{ Streams} & -    &- & - & Gaus.  & - & Gaus. & Median & Gaus.+Median   \\
			\textbf{Perturbation} & & & &  \\
			\hline
			Original(seen) & 0.43 &0.43&0.49&0.43 & 0.48 & \textbf{0.52} &0.46 &0.50\\
			Gaussian &  0.12&0.17&0.17&0.14&0.14&0.24&\textbf{0.28}&\textbf{0.28}\\
			Shot & 0.22&0.24&0.27&0.22&0.24&\textbf{0.38}&0.31&0.34\\
			Impulse & 0.33& 0.35 & 0.39 & 0.34 & 0.38 & 0.45 & 0.46 & \textbf{0.48} \\
			\hline
		\end{tabular}
	\end{center}
	\label{Tab:detection}
\end{table*}

\subsubsection{Ablation: Relevance of the Different Streams}
We evaluated the importance of different streams in GIT. For this purpose we tested different combinations in Tab.~\ref{Tab:Streams}. In the first two columns we only used the Gaussian (Gaus.) and the autoencoder (AE) and gradually added the others. The more streams GIT considers, the better the detection performance of the method irrespective of the ordering of the added streams. The method is able to decide which stream delivers the most relevant information automatically which further justifies the multistream concept.

\subsubsection{Ablation: Relevance of the seen perturbations}
\label{Abl.Perturbations}
In our main experiment shown in Tab.~\ref{Tab:Final}, we used only data from the Adversarial Example method FGSM to train GIT (and the other trainable detectors). The experiments show that using Adversarial Example data, GIT can generalize to other classes of perturbations like Predictive Uncertainty and Out-of-Distribution (albeit with mixed results in the case of Out-of-Distribution detection). This training procedure is similar to the experimental setups currently usual in the state-of-the-art and provides a glimpse on the generalization capabilities of the detectors over various perturbations. However, in practice one would train on all available perturbations to achieve a detector with best possible performance. Therefore, we further investigated the relevance of the variability of the seen perturbation training set by further training GIT on only original data and all perturbation datasets as shown in Tab.~\ref{Tab:Data}. The original data is not enough to cover the whole problem space and as expected the overall performance rises as more perturbation types are seen during training. Conversely, the variability in the adversarial FGSM data seems to already cover the other Adversarial Examples and Predictive Uncertainty setups since the results on these setups do not alter much when training with all perturbations. However, some improvement can be achieved for Out-of-Distribution data when all perturbations are seen during training. We conclude that, although generalization between different perturbation setups works, when seeking a best possible detector it is important to include at least some perturbations from each category.

\subsection{Extension to Object Detection}
We extended the method GIT to be applicable in the field of Object Detection to demonstrate the simple adaptability of GIT to other vision based problems.

\textbf{Necessary adaptations:} The output of an object detection problem is an undefined number of objects per image, each provided with classification and location information. The extraction of the gradients is based on the classification part of the network and all possible object candidates per image are considered individually. Similar as for image classification the gradients are computed using the classification loss function that receives as target the one hot vector of the predicted class of the corresponding object candidates. Then, the training of the head and the application of the invariance transformations is directly adopted from the classification case.

\textbf{ Experiments:} We based our experiments on an efficient Single-Shot-Detector (SSD) \cite{liu2016ssd} evaluated on KITTI \cite{geiger2013vision} which was split into 80\% training data for the SSD, 10\% training data for detection method and 10\% evaluation data, implementation details are provided in the Appendix.

%

The evaluation of uncertainty or error prediction in the case of object detection is not as straightforward as in classification. When applying the uncertainty measures only on the final output of the SSD, and hence the actually detected objects, false negatives are not considered. Therefore, we evaluate the methods using the mean Average Precision (mAP) that directly considers the predicted uncertainty of the methods: Each detection method predicts uncertainty information which can be directly used as object confidence. The classification-score of the predicted class is replaced with the predicted object confidence. Consequently, a better confidence estimate results in an improved mAP score. A more detailed explanation can be found in the Appendix. 

As baseline methods we used Monte Carlo Dropout \cite{miller2018dropout}, Deep Ensembles, invariance transformations and gradient information \cite{riedlinger2021gradient} on their own. Implementation details on the methods can be found in the Appendix. Each method has only the original data on hand during training and needs to generalize to the perturbed data. Results are shown in Table~\ref{Tab:detection}.

In most cases the multistream approach is able to cover the performance of the best single stream or even improves it. Gaussian transformation and gradient information each on its own are unable to outperform the baselines given by MC-Dropout and DeepEnsemble. However, their combination outperforms the other methods even on the non perturbed original setup which further shows the good interaction of the gradients and invariance transformations and the applicability of GIT beyond classification. Consequently, GIT can effectively be used to increase the robustness of object classifiers.

\section{Conclusion}
Most state-of-the-art error detection methods for DNNs focus only on a single reason for misclassification. However, in real-world applications the reasons for misclassification are often unknown and diverse and therefore, generalization to a wide variety of perturbations is necessary. Furthermore, current approaches do not consider perturbed samples that are still correctly classified. They either ignore their existence or even assign them to the negative (misclassified) class. 

We therefore developed and investigated a novel detection method that combines \textbf{G}radient information and \textbf{I}nvariance \textbf{T}ransformations (GIT) in a multistream approach and built up an extensive experimental set-up to cover the detection of Out-of-Distribution, Predictive Uncertainty and Adversarial Examples on several datasets and network architectures. While GIT was trained on only a single perturbation type, we evaluated the generalization capability to other perturbation types. The experiments show that the multistream concept leads GIT to a robust detection of misclassified samples of all types. GIT is on par with state-of-the art for Out-of-Distribution detection and highly superior for other reasons of generalization failures, especially in difficult situations when the perturbed data points lie close to the decision boundary. We further examined GIT in ablations studies and demonstrated that GIT can be easily extended to object detection, which paves the way towards more application fields. 

In future work, we want to evaluate more sophisticated transformation streams such as Generative Adversarial Networks or Variational Autoencoders and investigate replacing the current stream-wise training with an end-to-end training. We aim to further extend and improve GIT for object detection and other application areas. Since the used invariances are currently based on prior knowledge of the image domain, other fields would require other, domain-specific invariances.

\bibliographystyle{plain}
\bibliography{mybibfile}

\section*{Appendix}

\subsubsection{Evaluations using TNR at 95\% TPR}
To have further insight into the performance of GIT we additionally used TNR at TPR 95\% as metric. Results are shown in Table \ref{Tab:TNR}.\\

\subsubsection{Hyper-Parameters for the Perturbation Generation}
The hyper-parameters (for FGSM and BIM the adversarial noise level; for DeepFool and CWL2 the step size; the noise factors for the generation of the gaussian, shot and impulse perturbations) for the generation of the perturbation set-ups for DenseNet and ResNet are given in Table \ref{Tab:PUD}. \\

\begin{table*}[t]
	\caption{TNR at TPR 95\% score for the detection methods Mahalanobis, GraN, GradNorm and GIT (own). \emph{(seen)} marks the perturbation set used for training the backbones of the trainable detectors.}
	\begin{center}
		\begin{tabular}{l l lc  c c c c}
			\hline
			\multirow{3}{*}{\rotatebox[origin=c]{90}{\textbf{Model}}}&&\multicolumn{1}{r}{\textbf{Data}} & \textbf{CIFAR-10}&\textbf{CIFAR-100}&\textbf{SVHN} \\
			&\multicolumn{2}{r}{\textbf{Detector}}& \multicolumn{4}{c}{\textbf{Mahalanobis / GraN / GradNorm / GIT (own)}}\\

			&\multicolumn{2}{l}{\textbf{Perturbation}}& & & &\\

			\hline
			\multirow{11}{*}{\rotatebox[origin=c]{90}{ResNet}}&\multirow{4}{*}{\rotatebox[origin=c]{90}{P.Unc.}}&Original &33.86/42.59/24.41/\textbf{53.54}& \textbf{39.86}/14.52/22.58/25.35& 35.84/\textbf{63.58}/21.39/44.51\\
			&&Gaussian &14.64/~8.53/20.36/\textbf{69.21}& 19.59/28.00/16.62/\textbf{61.44}& 22.07/47.52/20.26/\textbf{82.89}\\
			&&Shot &    19.01/~9.69/14.79/\textbf{66.20} & 26.43/20.92/15.82/\textbf{91.43} &  20.15/49.54/21.00/\textbf{80.36}\\
			&&Impulse &  33.90/24.55/15.59/\textbf{83.20} & 29.34/20.50/21.45/\textbf{54.47} &22.47/35.24/21.81/\textbf{87.27}\\
			\cline{2-7}
			&\multirow{4}{*}{\rotatebox[origin=c]{90}{Adv.}}&FGSM(seen) & 25.53/50.05/15.20/\textbf{86.42}& 36.68/36.17/31.81/\textbf{65.25}& 21.95/35.27/17.28/\textbf{70.92}\\
			&&BIM  &  32.35/64.60/18.51/\textbf{90.64}& 32.73/39.74/39.24/\textbf{62.06}&  40.67/41.56/19.81/\textbf{85.51} \\
			&&DeepFool&  ~4.03/32.33/19.23/\textbf{70.39}& ~5.70/11.90/~5.29/\textbf{22.99}& ~3.00/56.73/~4.94/\textbf{75.88}\\
			&&CWL2 &  36.77/69.74/14.43/\textbf{93.89} & 50.05/45.88/13.84/\textbf{66.53}& 45.45/84.53/11.49/\textbf{91.36}\\
			\cline{2-7}
			&\multirow{3}{*}{\rotatebox[origin=c]{90}{OOD}}&SVHN/C/Nat   &\textbf{88.64}/~5.70/17.97/~8.31 & \textbf{25.71}/14.21/13.81/17.99 &\textbf{84.99}/81.01/33.59/53.89\\
			&&ImgN/Places &21.64/21.21/33.69/\textbf{40.49}&\textbf{20.75}/11.61/19.00/~9.60&90.39/\textbf{86.15}/34.17/56.03\\
			&&LSUN/SUN & 23.86/24.85/35.58/\textbf{40.65} & 22.50/11.19/17.84/\textbf{79.52} & \textbf{87.84}/85.38/32.79/49.20  \\

			\hline
			\multirow{11}{*}{\rotatebox[origin=c]{90}{DenseNet}}&\multirow{4}{*}{\rotatebox[origin=c]{90}{P.Unc.}}&Original &~4.12/20.62/\textbf{29.90}/19.59& 12.95/\textbf{23.44}/8.26/22.77 &42.86/\textbf{61.90}/13.23/32.80 \\
			&&Gaussian &~7.49/31.28/21.05/\textbf{82.69} & ~4.06/35.75/10.14/\textbf{48.24} & 10.93/42.59/16.83/\textbf{86.32}\\
			&&Shot & ~8.06/21.45/19.64/\textbf{73.92} & ~3.37/34.49/13.27/\textbf{48.98} & 10.59/38.58/11.40/\textbf{84.80}\\
			&&Impulse & ~7.00/30.38/12.36/\textbf{83.11} & ~4.44/33.64/11.46/\textbf{89.29} & 10.12/30.25/15.30/\textbf{80.89}\\
			\cline{2-6}
			&\multirow{4}{*}{\rotatebox[origin=c]{90}{Adv.}}&FGSM(seen) &~5.17/54.81/~6.18/\textbf{90.37} & ~5.87/47.17/~6.17/\textbf{61.03} & 16.34/34.39/11.24/\textbf{73.69} \\
			&&BIM & 10.71/67.58/~4.55/\textbf{92.63}& ~6.39/51.32/~6.90/\textbf{63.08}&  48.22/67.67/52.91/\textbf{89.46} \\
			&&DeepFool &~4.72/27.51/~5.22/\textbf{56.12} & ~5.70/20.00/~6.30/\textbf{25.70} & 15.74/\textbf{86.76}/15.74/82.04 \\
			&&CWL2 &   ~9.22/74.57/11.55/\textbf{93.92}& ~4.60/63.60/~9.40/\textbf{68.00} & 40.69/93.30/18.31/\textbf{93.81} \\
			\cline{2-6}
			&\multirow{3}{*}{\rotatebox[origin=c]{90}{OOD}}&SVHN/C & \textbf{67.96}/23.13/48.14/12.92 & ~5.43/13.86/\textbf{65.57}/~9.00&63.29/\textbf{74.98}/17.03/60.60   \\
			&&ImgN & ~0.60/16.71/\textbf{78.99}/16.40 &~1.58/14.02/10.78/~0.56&60.50/\textbf{84.69}/47.88/62.60\\
			&&LSUN &  ~0.19/16.25/\textbf{83.81}/20.35&  ~1.64/\textbf{17.27}/~6.70/~0.45 & 62.00/\textbf{82.46}/51.82/57.71 \\
			\cline{1-6}
		\end{tabular}		
	\end{center}
	\vspace{-0pt}
	\label{Tab:TNR}
\end{table*}

\begin{table}[t]
	\caption{Perturbation generation hyper-parameters DenseNet.}
	\begin{center}
		\begin{tabular}{ llllll}
			\hline
			&Method  & CIFAR10 & CIFAR100 & SVHN &ImgNet\\
			\hline
			\multirow{7}{*}{\rotatebox[origin=c]{90}{ResNet}}&Gaussian & 0.0057&0.00155&0.03&0.001\\
			&Shot &70&300&13&300\\
			&Impulse &0.025&0.011&0.1&0.0001\\
			\cline{2-6}
			&FGSM & 0.0065&0.002&0.05&0.0005\\
			&BIM &0.0007&0.00032&0.0022&0.000001\\
			&DeepFool &0.084&0.00975&0.085&-\\
			&CWL2 & 0.00014&0.000075&0.000378& 0.0001\\
			\hline	
			\multirow{7}{*}{\rotatebox[origin=c]{90}{DenseNet}}&Gaussian & 0.003&0.0008&0.028&0.0001\\
			&Shot &120&600&15&300\\
			&Impulse &0.032&0.012&0.11&0.0001\\
			\cline{2-6}
			&FGSM & 0.004&0.0012&0.06&0.0005\\
			&BIM &0.0005&0.0002&0.00135& 0.000001\\
			&DeepFool &0.10002&0.009995&0.1001&-\\
			&CWL2  & 0.000095&0.00004& 0.000245&0.000015\\
			\hline	
		\end{tabular}
	\end{center}
	\vspace{-0pt}
	\label{Tab:PUD}
\end{table}

\subsubsection{Classification DNNs}
We based our experiments on DenseNets \cite{huang2017densely} and ResNets \cite{he2016deep}. The experiments for the datasets Cifar10, Cifar100 and SVHN follow the same setup as in \cite{lee2018simple}. We used their pretrained DenseNet with 100 layers and their pretrained ResNet with 34 layers. Their models are available at \url{https://github.com/pokaxpoka/deep_Mahalanobis_detector}. Our ImageNet Experiments are adapted from \cite{huang2021importance}. For ImageNet we used a ResNetv2-101 and a DenseNet-
121 model. A trained version is available at \url{https://github.com/google-research/big_transfer}.\\

\subsubsection{Autoencoder Stream}
The architecture of the Autoencoder used for one stream of GIT is shown in Figure \ref{Tab:AE}. For each dataset the Autoencoder is trained in a self-supervised manner on the corresponding original training dataset using the BCE-Loss and Adam as optimizer.\\

\begin{table}[h]
	
	\caption{Architecture of the Autoencoder}
	\begin{center}
		\begin{tabular}{ccccccc}
			\hline
			Module & $c_{in}$ & $c_{out}$ & K & Str. & Pad. & Act. \\
			\hline
			Conv2d & 3 & 12 & 4$\times$4 & 2 & 1 & ReLU\\
			Conv2d & 12 & 24 & 4$\times$4 & 2 & 1 & ReLU\\
			Conv2d & 24 & 48 & 4$\times$4 & 2 & 1 & ReLU\\
			ConvTransp2d& 48 &24 & 4 & 2 &1 &ReLU\\
			ConvTransp2d& 24 &12 & 4 & 2 &1 &ReLU\\
			ConvTransp2d& 12 &3 & 4 & 2 &1 &Sigmoid\\
			\hline
		\end{tabular}
	\end{center}
	\vspace{-0pt}
	\label{Tab:AE}
\end{table}

\subsubsection{GradNorm using all layers}
Huang et al. \cite{huang2021importance} introduced the detection method GradNorm based on the norm of the gradients of the last layer of the neural network. We adapted GradNorm (ad.GradNorm) by using the gradient information of all layers and trained a logistic regression network to combine the gradient features, similarly to the other data based detectors. Similarly we adapted GIT to only be based on the last layer. Results for GIT are quite similar except for some Out-of-Distribution setups where the method in which all layers are used leads to a higher score. For GradNorm the results improve when all layers are used for the adversarial case but get worse for the others. Table \ref{Tab:gradnorm} exemplary shows the results for CIFAR-10 on DenseNet. In general, GIT outperforms GradNorm, even when all layers are considered.\\

\begin{table}[h]
	\caption{AUROC scores for GIT, GradNorm and adapted GradNorm on DenseNet for CIFAR-10.}
	\begin{center}
		\begin{tabular}{c l|c c c c }
			\hline
			&\multicolumn{1}{r|}{Method} &  ad.~GIT   &GIT  &GradNorm &ad.~GradNorm \\
			&\multicolumn{1}{r|}{Used layers} &  last  &all  &last&all \\
			\hline
			\multirow{4}{*}{\rotatebox[origin=c]{90}{P.Unc.}}&Original &87.28 &87.22&78.29 &79.56\\
			&Gaussian & 96.36 &96.22&72.24&65.81\\
			&Shot & 95.38 &95.09 &75.04& 65.71\\
			&Impulse & 96.48 & 96.00&66.33&79.28 \\
			\hline
			\multirow{4}{*}{\rotatebox[origin=c]{90}{Adv.}}&FGSM(seen) & 97.27 &97.50&55.39& 86.05\\
			&BIM & 98.09 &98.42&54.97&87.04 \\
			&DeepFool & 91.61 &91.82&51.35&60.84\\
			&CWL2 & 98.46 &98.76 &64.00&85.77 \\
			\hline
			\multirow{3}{*}{\rotatebox[origin=c]{90}{OOD}}&SVHN & 69.83 &73.87&84.75&90.90 \\
			&ImageNet & 83.67 &83.66&95.29&73.56\\
			&LSUN &85.89 &85.37 &96.96&73.74\\
			\hline
		\end{tabular}
		\vspace{-0pt}
	\end{center}
	
	\label{Tab:gradnorm}
\end{table}

\subsubsection{Object Detection}
The SSD's network architecture and hyper-parameters are taken from \url{https://github.com/amdegroot/ssd.pytorch}. As weight initialisation we used Kaiming-Normal due to a small performance improvement. 

For the Monte Carlo Dropout variant, two dropout layers were added to the last two
convolutional layers of the SSD’s backbone. The hyper-parameters are adopted as described in reference \cite{miller2018dropout}. Dropout is used during both the training
and the testing phase. During testing 10 forward runs are conducted in order to sample different predictions.\\
For the Deep Ensemble method the 10 different predictions were samples by differently trained networks, each trained from another set of randomly initialised weights.
Both, the results of MC-Dropout and of the Deep Ensemble method are merged using the intersection over union based approach proposed in reference \cite{miller2019evaluating}. The merging threshold is set to $0.7$ and only boxes of the same class are merged.

Gaussian perturbations were sampled from a Gaussian distribution with a standard deviation of $10.0$. For Gaussian smoothing the standard deviation is chosen from the set $\lbrace 0.1, 0.2, \dots, 1.0\rbrace$. For the Median filter the size of the filter is chosen from $\lbrace2\times2, 3\times3,\dots,10\times10\rbrace$. Both hyper-parameters are optimised using the non-perturbed validation data. 

In the object detection setup, we directly use the mAP score in order to evaluate the uncertainty scores. Usually, the highest classification score output is used to accept or reject candidate bounding boxes (e.g. via non-maximum suppression and thresholding) and hence directly influences the mAP scores. By replacing this highest score with a confidence measure provided by the detection methods, a higher mAP can be achieved, e.g. by correctly rejecting false positive boxes. The better the confidence measure, the higher the mAP score.

\end{document}